\documentclass[twocolumn]{article}
\usepackage{graphicx}
\usepackage[fleqn]{amsmath}
\usepackage[psamsfonts]{amssymb}
\usepackage{multirow}
\usepackage{algorithmic}
\usepackage{algorithm}
\usepackage{printlen}

\graphicspath{{./figures/}}

\newlength\myindent
\title{Reversible designs for extreme memory cost reduction of CNN training}

\author{%
	\textsc{Tristan Hascoet}\thanks{Equal contribution} \thanks{Corresponding author} \\ 
	\normalsize Kobe Univeristy \\ 
	\normalsize tristan@people.kobe-u.ac.jp 
	\and 
	\textsc{Quentin Febvre}\footnotemark[1] \\ 
	\normalsize Kobe Univeristy \\ 
    \normalsize Sicara \\ 
    \normalsize quentin.febvre@gmail.com  
    \and 
    \textsc{Yasuo Ariki} \\ 
    \normalsize Kobe Univeristy \\ 
	\normalsize ariki@kobe-u.ac.jp 
	\and 
    \textsc{Tetsuya Takiguchi} \\ 
    \normalsize Kobe Univeristy \\ 
	\normalsize takigu@kobe-u.ac.jp 
}

\date{\vspace{-1ex}}

\begin{document}
	
\maketitle

\begin{abstract} 
Training Convolutional Neural Networks (CNN) is a resource intensive task that requires specialized hardware for efficient computation. 
One of the most limiting bottleneck of CNN training is the memory cost associated with storing the activation values of hidden layers needed for the computation of the weights gradient during the backward pass of the backpropagation algorithm.
Recently, reversible architectures have been proposed to reduce the memory cost of training large CNN by reconstructing the input activation values of hidden layers from their output during the backward pass, circumventing the need to accumulate these activations in memory during the forward pass.
In this paper, we push this idea to the extreme and analyze reversible network designs yielding minimal training memory footprint.
We investigate the propagation of numerical errors in long chains of invertible operations and analyze their effect on training.
We introduce the notion of pixel-wise memory cost to characterize the memory footprint of model training,
and propose a new model architecture able to efficiently train arbitrarily deep neural networks with a minimum memory cost of 352 bytes per input pixel.
This new kind of architecture enables training large neural networks on very limited memory, opening the door for neural network training on embedded devices or non-specialized hardware. For instance, we demonstrate training of our model to 93.3\% accuracy on the CIFAR10 dataset within 67 minutes on a low-end Nvidia GTX750 GPU with only 1GB of memory.
\end{abstract}




\section{Introduction}
Over the last few years, Convolutional Neural Networks (CNN) have enabled unprecedented progress on a wide array of computer vision tasks.
One disadvantage of these approaches is their resource consumption: 
Training deep models within a reasonable amount of time requires special
Graphical Processing Units (GPU) with numerous cores and large memory capacity.
Given the practical importance of these models, a lot of research effort has been directed
towards algorithmic and hardware innovations to improve their resource efficiency such as low-precision arithmetic \cite{jacob2018quantization}, network pruning for inference \cite{molchanov2016pruning}, or efficient stochastic optimization algorithms \cite{kingma2014adam}.

In this paper, we focus on a particular aspect of resource efficiency: optimizing the memory cost of training CNNs. 
We envision several potential benefits from the ability to train large neural networks within limited memory:

\textbf{Democratization of Deep Learning research:} 
Training large CNN requires special GPUs with large memory capacity. 
Typical desktop GPUs memory capacity is too small for training large CNNs.
As a result, getting into deep learning research comes with the barrier cost of either buying specialized hardware or renting live instances from cloud service providers. 
Reducing the memory cost of deep model training would allow training deep nets on standard graphic cards without the need for specialized hardware, effectively removing this barrier cost.
In this paper, we demonstrate efficient training of a CNN on the CIFAR10 dataset (93.3\% accuracy within 67 minutes) on an Nvidia GTX750 with only 1GB of memory.

\textbf{On-device training:}
With mobile applications, a lot of attention has been given to optimize inference on edge devices with limited computation resources.
Training state-of-the-art CNN on embedded devices, however, has still received little attention.
Efficient on-device training is a challenging task for the underlying power efficiency, computation and memory optimization challenges it involves.
As such, CNN training has thus far been relegated to large cloud servers, and trained CNNs are typically deployed to embedded device fleets over the network.
On-device training would allow bypassing these server-client interactions over the network.
We can think of several potential applications of on-device training, including:
\begin{itemize}
 \item Life-long learning: Autonomous systems deployed in evolving environments like drones, robots or sensor networks might benefit from continuous life-long learning to adapt to their changing environment.
On-device training would enable such application without the expensive communication burden of having edge devices continuously sending their data to remote servers over the network. It would also provide resilience to network failures in critical application scenarios.
 \item In privacy-critical applications such as biometric mobile phone authentication, users might not want to have their data sent over the network. 
On-device training would allow fine-tuning recognition models on local data without sending sensitive data over the network.
\end{itemize}
In this work, we propose an architecture with minimal training memory cost requirements which enables training within the tight memory constraints of embedded devices. 

\textbf{Research in optimization:}
Recent works on stochastic optimization algorithms have highlighted the benefits of large batch training \cite{shallue2018measuring,largebatch}.
For example, in Imagenet, linear speed-ups in training have been observed with increasing batch sizes up to tens of thousands of samples \cite{largebatch}.
Optimizing the memory cost of CNN training may allow further research on the optimization trade-offs of large batch training.
For small datasets like MNIST or CIFAR10, we are able to process the full dataset in 14 and 18 GB of memory respectively.
Although large batch training on such small dataset is very computationally inefficient with current stochastic optimization algorithms \cite{largebatch},
the ability to process the full dataset in one pass allows to easily train CNNs on the true gradient of the error.
Memory optimization techniques have the potential to facilitate research on optimization techniques outside the realm of Stochastic Gradient Descent to be investigated.

In this paper, we build on recent works on reversible networks \cite{gomez2017reversible,jacobsen2018revnet} and ask the question: 
how far can we reduce CNN training memory cost using reversible designs with minimal impact on the accuracy and computational cost?
To do so, we take as a starting point the Resnet-18 architecture and analyze its training memory requirements.
We then analyze the memory cost reduction of invertible designs successively introduced in the RevNet and iRevNet architectures.
We identify the memory bottleneck of such architectures, which leads us to introduce a layer-wise invertible architecture.
However, we observe that layer-wise invertible networks accumulate numerical errors across their layers, which leads to numerical instabilities impacting model accuracy.
We characterize the accumulation of numerical errors within long chains of revertible operations and investigate their effect on model accuracy.
To mitigate the impact of these numerical errors on the model accuracy, we propose both a reparameterization of invertible layers and a hybrid architecture combining the benefits of layer-wise and residual-block-wise reversibility to stabilize training.

Our main result is to present a new architecture that allows to efficiently train a CNN with the minimal memory cost of 352 bytes per pixel.
We demonstrate the efficiency of our method by efficiently training a model to 93.3\% accuracy on the CIFAR10 dataset within 67 minutes on a low-end Nvidia GTX750 with only 1GB of VRAM.

\section{Related Work}
\subsection{Reversibility}

Reversible network designs have been proposed for various purposes including generative modeling, visualization, solving inverse problems, or theoretical analysis of hidden representations.

Flow-based generative models use analytically invertible transformations to compute the change of variable formula. Invertibility is either achieved through channel partitioning schemes (NICE \cite{dinh2014nice} Real-NVP \cite{dinh2016density}), weight matrix factorization (GLOW \cite{kingma2018glow}) or constraining layer architectures to easily invertible unitary operations (Normalization flows \cite{rezende2015variational})

Neural ODEs \cite{chen2018neural} take a drastically different take on invertibility: They leverage the analogy between residual networks and the Euler method to define continuous hidden state systems.
The conceptual shift from a finite set of discrete transformations to a continuous regime gives them invertibility for free. 
The computational efficiency of this approach, however, remains to be demonstrated.

The RevNet model \cite{gomez2017reversible} was inspired by the Real-NVP generative model. They adapt the idea of channel partitioning and propose an efficient architecture for discriminative learning.
The iRevNet \cite{jacobsen2018revnet} model builds on the RevNet architecture: they propose to replace the irreversible max-pooling operation with an invertible operation that reshapes the hidden activation states
so as to compensate the loss of spatial resolution by an increase in the channel dimension.
By preserving the volume of activations, their pooling operation allows for exact reconstruction of the inverse.
In their original work, the authors focus on the analysis of the representations learned by invertible models rather than resource efficiency.
From a resource optimization point of view, one downside of their method is that the proposed invertible pooling scheme drastically increases the number of channels in upper layers.
As the size of the convolution kernel weights grows quadratically in the number of channels, the memory cost associated with storing the model weights becomes a major memory bottleneck.
We address this issue in our proposed architecture.
In \cite{jacobsen2018excessive}, the authors use these reversible architectures to study undesirable invariances in feature space.

In \cite{behrmann2018invertible}, the authors propose a unified architecture performing well on both generative and discriminative tasks.
They enforce invertibility by regularizing the weights of residual blocks so as to guarantee the existence of an inverse operation.
However, the computation of the inverse operation is performed with power iteration methods which are not optimal from a computational perspective.

Finally, \cite{rota2018place} propose to reconstruct the input activations of normalization and activation layers using their inverse function during the backward pass.
We propose a similar method for layer-wise invertible networks. 
However, as their model does not invert convolution layers, 
it does not feature long chains of invertible operations so that they do not  need to account for numerical instabilities.
Instead, our proposed model features long chains of invertible operations so that we need to characterize numerical errors in order to stabilize training.

\subsection{Resource efficiency}

Research into resource optimization of CNNs covers a wide array of techniques, most of which are orthogonal to our work. We briefly present some of these works:

On the architectural side, Squeezenet \cite{iandola2016squeezenet} was first proposed as an efficient neural architecture reducing the number of model parameters while maintaining high classification accuracy.
MobileNet \cite{howard2017mobilenets} uses depth-wise separable convolutions to further reduce the computational cost of inference for embedded device applications.

Network pruning \cite{molchanov2016pruning} is a set of techniques developed to decrease the model weight size and computational complexity.
Network pruning works by removing the network weights that contribute the least to the model output.
Pruning deep models has been shown to drastically reduce the memory cost and computational cost of inference without  significantly hurting model accuracy.
Although pruning has been concerned with optimization of the resource inference, the recently proposed lottery ticket hypothesis \cite{frankle2018lottery} has shown that specifically pruned networks could  be trained from scratch to high accuracy.
This may be an interesting and complementary line of work to investigate in the future to reduce training memory costs.

Low precision arithmetic has been proposed as a mean to reduce both memory consumption and computation time of deep learning models. Mixed precision training \cite{micikevicius2017mixed} combines float16 with float32 operations to avoid numerical instabilities due to either overflow or underflow.
For inference,  integer quantization \cite{jacob2018quantization,wu2018training} has been shown to drastically improve the computation and memory efficiency and has been successfully deployed on both edge devices and data centers.
Integrating mixed-precision training to our proposed architecture would allow us to further reduce training memory costs. 

Most related to our work, gradient checkpointing was introduced as a mean to reduce the memory cost of deep neural network training.
Gradient checkpointing, first introduced in \cite{martens2012training}, trades off memory for computational complexity by storing only a subset of the activations during the forward pass.
During the backward pass, missing activations are recomputed from the stored activations as needed by the backpropagation algorithm.
Follow-up work \cite{chen2016training} has since built on the original gradient checkpointing algorithm to improve this memory/computation trade-off.  
However, reversible models like RevNet have been shown to offer better computational complexity than gradient checkpointing,
at the cost of constraining the model architecture to invertible residual blocks.

\section{Preliminaries}

In this section, we analyze the memory footprint of training architectures with different reversibility patterns.
We start by introducing some notations and briefly review the backpropagation algorithm
in order to characterize the training memory consumption of deep neural networks. 
In our analysis, we use a Resnet-18 as a reference baseline and analyze its training memory footprint.
We then gradually augment the baseline architecture with reversible designs and analyze their impact on computation and memory consumption.

\subsection{Backpropagation \& Notations}

Let us consider a model $F$ made of $N$ sequential layers trained to minimize the error $e$ defined by a loss function $\mathcal{L}$ for an input $x$ and ground-truth label $\bar{y}$:

 \begin{subequations}
 	\begin{align}
 	F &: x \rightarrow y \\
 	y &= f_N \circ ... \circ f_2 \circ f_1(x) \\
 	e &=  \mathcal{L}(y, \bar{y})
 	\end{align}
 \end{subequations}

During the forward pass, each layer $f_i$ takes as input the activations $z_{i-1}$ from the previous layer and outputs activation features $z_i=f_i(z_{i-1})$, with $z_0=x$ and $z_N=y$ being the input and output of the network respectively.

During the backward pass, the gradient of the loss with respect to the hidden activations are propagated backward through the layers of the networks using the chain rule as:


\begin{equation}
\frac{\delta \mathcal{L}}{\delta z_{i-1}} = \frac{\delta \mathcal{L}}{\delta z_{i}}  \times \frac{\delta z_{i}}{\delta z_{i-1}}
\end{equation}

Before propagating the loss gradient with respect to its input to the previous layer, 
each parameterized layer computes the gradient of the loss with respect to its parameters. 
In vanilla SGD, for a given learning rate $\eta$, the weight gradients are subsequently used to update the weight values as:

\begin{subequations}
\begin{align}
\frac{\delta \mathcal{L}}{\delta \theta_i} & =\frac{\delta \mathcal{L}}{\delta z_{i}}  \times \frac{\delta z_{i}}{\delta \theta_i} \\
\theta_i & \leftarrow \theta_i - \eta \times \frac{\delta \mathcal{L}}{\delta \theta_i}
\end{align}
\end{subequations}

However, the analytical form of the weight gradients are functions of the layer's input activations $z_{i-1}$. 
In convolution layers, for instance, the weight gradients can be computed as the convolution of the input activation by the output's gradient:

 \begin{equation}
\frac{\delta \mathcal{L}}{\delta \theta_i} = z_{i-1} \star \frac{\delta \mathcal{L}}{\delta z_i} 
 \end{equation}

Hence, computing the derivative of the loss with respect to each layer's parameters $\theta_i$ requires knowledge of the input activation values $z_{i-1}$.  
In the standard backpropagation algorithm, hidden layers activations are stored in memory upon computation during the forward pass. 
Activations accumulate in live memory buffers until used for the weight gradients computation in the backward pass. 
Once the weight gradients computed in the backward pass, the hidden activation buffers can be freed from live memory. 
However, the accumulation of activation values stored within each parameterized layer along the forward pass creates a major bottleneck in GPU memory.

The idea behind reversible designs is to constrain the network architecture to feature invertible transformations.
Doing so, activations $z_i$ in lower layers can be recomputed through inverse operations from the activations $z_{j>i}$ of higher layers.
In such architectures, activation do not need to be kept in memory during the forward pass as they can be recomputed from higher layer activations during the backward pass, effectively freeing up the GPU live memory.

\subsection{Memory footprint}

We denote the memory footprint of training a neural network as a value $\mathcal{M}$ in bytes. 
Given an input $x$ and ground truth label $\bar{y}$, the memory footprint represents the peak memory consumption during an iteration of training including the forward and backward pass.
We divide the total training memory footprint $\mathcal{M}$ into several memory cost factors: the cost $M_{\theta}$ of storing the model weights, the hidden activations  $M_{z}$, and the gradients $M_{g}$:

\begin{equation}
\mathcal{M} = M_{\theta} + M_{z} + M_{g}
\end{equation}

In the following subsections, we detail the memory footprint of existing architectures with different reversibility patterns.
To help us formalize these memory costs, we further introduce the following notations: 
let $n(x)$ denote the number of elements in a tensor $x$, i.e.; if $x$ is an $h \times w$ matrix, then $n(x)=h \times w$. 
Let $bpe$ be the memory cost in bytes per elements of a given precision so that the actual memory cost for storing an $h \times w$ matrix is $n(x) \times bpe$. 
For instance, float32 tensors have a memory cost per element $bpe=4$. 
We use $bs$ to denote the batch size, and $c_i$ to denote the number of channels at layer $i$.

\subsection{Vanilla ResNet}

The architecture of a vanilla ResNet-18 is shown in Figure 1.
Vanilla ResNets do not use reversible computations so that the input activations of all parameterized layers need to be accumulated in memory during the forward pass for the computation of the weight gradients to be done in the backward pass.

Hence the peak memory footprint of training a vanilla ResNet happens at the beginning of the backward pass when the top layer's activation gradients need to be stored in memory in addition to the full stack of hidden activation values.

Let us denote by $P \subset N$ the subset of parameterized layers of a network $F$ 
(i.e.; convolutions and batch normalization layers, excluding activation functions and pooling layers). 
The memory cost associated with storing the hidden activation values is given by: 

\begin{subequations}
\begin{align}
M_{z} &= \sum_{i \in P} n(z_i) \times bpe \\
      &= \sum_{i \in P} bs \times c_i \times h_i \times w_i \times bpe 
\end{align}
\end{subequations}

Where $h_i$ and $w_i$ represent the spatial dimensions of the activation values at layer $i$.
$h_i$ and $w_i$ are determined by the input image size $h \times w$ and the pooling factor $p_i$ of layer $i$, so we can factor out both the spatial dimensions and the batch size from this equation, yielding a memory cost per input pixel $M'_z$:

\begin{subequations}
\begin{align}
M_{z} &= \sum_{i \in P} bs \times h \times w \times p_i \times c_i \times bpe \\
      &= bs \times h \times w \times \sum_{i \in P} p_i \times c_i \times bpe \\
M_{z}' &= \frac{M_{z}}{bs \times h \times w}  \\
       &= \sum_{i \in P} p_i \times c_i \times bpe 
\end{align}
\end{subequations}

The memory footprint of the weights is given by:

\begin{equation}
 M_{\theta} = \sum_{i \in P}  n(\theta_i)\times bpe 
\end{equation}

The memory footprint of the gradients correspond to the size of the gradient buffers at the time of peak memory usage. In a vanilla ResNet18 model, this peak memory usage happens during the backward pass through the last convolution of the network.
Hence, the memory footprint of the gradients correspond to the memory cost of storing the gradients with respect to either the input or the output of this layer, which also depends on the input pixel size:

\begin{subequations}
\begin{align}
M_{g}  &= max(n(g_{i-1}), n(g_i)) \times bpe \\
       &= h \times w \times bs \times p_i \times max(c_{i-1}, c_i) \times bpe\\
M_{g}' &= p_i \times max(c_{i-1}, c_i) \times bpe
\end{align}
\end{subequations}

Figure 1 illustrates the peak memory consumption of a ResNet-like architecture.
For a ResNet parameterized following Table 1, the peak memory consumption can then be computed as:

\begin{subequations}
\begin{align}
\mathcal{M} &= M_{\theta} + M_{z} + M_{g} \\
            &= M_{\theta} + (M_{z}' + M_{g}') \times (h \times w \times bs) \\
            &= 12.5*10^6 + 1928 \times (h \times w \times bs) \\ 
\end{align}
\end{subequations}

For example, a training iteration over a typical batch of 32 images of resolution $240 \times 240$ 
requires 12.5 MB of memory to store the model weights and 3.8 GB of memory to store the hidden layers activations and gradients
for a total of $\mathcal{M}=3.81$ GB of VRAM.
The memory cost of the hidden activations is thus the main memory bottleneck of CNN training as the cost associated with 
the model weights is negligible in comparison.

\subsection{RevNet}
The RevNet architecture introduces reversible blocks as drop-in replacements of the residual blocks of the ResNet architecture.
Reversible blocks have analytical inverses that allow for the computation of both their input and hidden activation values from the value of their output activations.
Two factors create memory bottlenecks in training RevNet architectures, which we refer to as the local and global bottlenecks.

First, the RevNet architecture features non-volume preserving max-pooling layers, for which the inverse cannot be computed.
As these layers do not have analytical inverses, their input must be stored in memory during the forward pass 
for the reconstruction of lower layer's activations to be computed during the backward pass. 
We refer to the memory cost associated with storing these activations as the global bottleneck, 
since these activations need to be accumulated during the forward pass through the full architecture.

The local memory bottleneck has to do with the synchronization of the reversible block computations:
While activations values are computed by a forward pass through the reversible block modules,
gradients computations flow backward through these modules so that the activations and gradient 
computations cannot be performed simultaneously.
Figure 2 illustrates the process of backpropagating through a reversible block:
First, the input activation values of the parameterized hidden layers within the reversible blocks are recomputed from the output.
Once the full set of activation have been computed and stored in GPU memory, the backpropagation of the gradients through the reversible block can begin.
We refer to the accumulation of the hidden activation values within the reversible block as the local memory bottleneck.

For a typical parameterization of a RevNet, as summarized in Table 1, 
the local bottleneck of lower layers actually outweighs the global memory bottleneck introduced by non-reversible pooling layers.
Indeed, as the spatial resolution decreases with pooling operations, 
the cost associated with storing the input activations of higher layers becomes 
negligible compared to the cost of storing activation values in lower layers.
Hence, surprisingly, the peak memory consumption of the RevNet architecture, as illustrated in Figure 3, 
happens in the backward pass through the first reversible block,
in which the local memory bottleneck is maximum.
For the architecture described in Table 1, the peak memory consumption can be computed as:

\begin{subequations}
\begin{align}
\mathcal{M} &= M_{\theta} + M_{z} + M_{g} \\
            &=(M_{\theta} + (M_z' + M_{g}') \times (h \times w \times bs) \\
            &= 12.7 \times 10^6 + 640 \times (h \times w \times bs) 
\end{align}
\end{subequations}

Following our previous example, a RevNet architecture closely mimicking the ResNet-18 architecture
requires $\mathcal{M}=1.19$ GB of VRAM for a training iteration over batch of 32 images of resolution $240 \times 240$.

Finally, the memory savings allowed by the reversible block come with the additional computational cost of computing the hidden activations during the backward pass.
As noted in the original paper, this computational cost is equivalent to performing one additional forward pass.

\subsection{iRevNet}

The iRevNet model builds on the RevNet architecture: they replace the irreversible max-pooling operation with an invertible operation that reshapes the hidden activation states
so as to compensate for the loss of spatial resolution by an increase in the channel dimension. 
As such, the iRevNet architecture is fully invertible, which alleviates the global memory bottleneck of the RevNet architecture.

This pooling operation works by stacking the neighboring elements of the pooling regions along the channel dimension,
i.e.; for a 2D pooling operation with $2 \times 2$ pooling window, the number of output channels is four times the number of input channels. 
Unfortunately, the size of a volume-preserving convolution kernel grows quadratically in the number of input channels:

\begin{subequations}
\begin{align}
M(\theta) &= c_{in} \times c_{out} \times k_h \times k_w \\
          &= c^2 \times k_h \times k_w
\end{align}
\end{subequations}

Consider an iRevNet network with initial channel size 32.
After three levels of $2 \times 2$ pooling, the effective channel size becomes $32 \times 4^3=2048$. A typical $3 \times 3$ convolution layer kernel for higher layers of such network would have $n(\theta)=2048^2 \times 3 \times 3=37M$ parameters.
At this point, the memory cost of the network weights $M_{\theta}$ becomes an additional memory bottleneck.

Furthermore, the iRevNet architecture does not address the local memory bottleneck of the reversible blocks.
Figure 4 illustrates such architecture. 
For an initial channel size of 32, as summarized in Table 1, the peak memory consumption is given by:

\begin{subequations}
\begin{align}
\mathcal{M} &= M_{\theta} + M_{z} + M_{g} \\
            &= M_{\theta} + (M_z' + M_{g}') \times (h \times w \times bs) \\
            &= 171 \times 10^6 + 640 \times (h \times w \times bs) 
\end{align}
\end{subequations}

Training such an architecture for an iteration over batches of 32 images of resolution $240 \times 240$ would require $\mathcal{M}=1.35$GB of VRAM.
In the next section, we introduce both layer-wise reversibility and a variant on this pooling operations to address the local memory bottleneck 
of reversible blocks and the weight memory bottleneck respectively.

\section{Method}

RevNet and iRevNet architectures implement reversible transformations at the level of residual blocks. 
As we have seen in the previous section, the design of these reversible blocks create a local memory bottleneck as all hidden activations within a reversible block need to be computed before the gradients are backpropagated through the block.
In order to circumvent this local bottleneck, we introduce layer-wise invertible operations. 
However, these invertible operations introduce numerical error, which we characterize in the following subsections.
In Section 5, we will show that these numerical errors lead to instabilities that degrade the model accuracy.
Hence, in section 4.2, we propose a hybrid model combining layer-wise and residual block-wise reversible operations to stabilize training while resolving the local memory bottleneck at the cost of a small additional computational cost.

\subsection{Layer-wise Invertibility}

In this section, we present invertible layers that act as drop-in replacement for convolution, batch normalization, pooling and non-linearity layers. We then characterize the numerical instabilities arising from the invertible batch normalization and non-linearities.

\subsubsection{Invertible batch normalization}

As batch normalization is not a bijective operation, it does admit an analytical inverse.
However, the inverse reconstruction of a batch normalization layer can be realized with minimal memory cost.
Given first and second order moment parameters $\beta$ and $\gamma$, the forward $f$ and inverse $f^{-1}$ operation of an invertible batch normalization layer can be computed as follows:

\begin{subequations}
\begin{align}
y = f(x) &= \gamma \times \frac{x - \hat{x}}{\sqrt{\dot{x}} + \epsilon} + \beta \\
x = f^{-1}(y, \hat{x}, \dot{x}) &= (\sqrt{\dot{x}} + \epsilon) \times \frac{y -  \beta}{\gamma}  + \hat{x}
\end{align}
\end{subequations}

Where $\hat{x}$ and $\dot{x}$ represent the mean and variance of $x$ respectively.
Hence, the input activation $x$ can be recovered from $y$ through $f^{-1}$ at the minimal memory cost of storing the input activation statistics $\hat{x}$ and $\dot{x}$.

Let us consider the accumulation of numerical errors arising from the inverse computation of an invertible batch normalization layer.
During the backward pass, the invertible batch norm layer is supposed to compute its input $x=f^{-1}(y, \hat{x}, \dot{x})$ from the output $y$.
In reality, however, the output recovered by upstream invertible layers is a noisy estimate $\hat{y}=y+\epsilon_y$ of the true output due to numerical errors introduced by upstream layers.
Let us define the signal to noise ratio (SNR) of the input and output signal as follows:

\begin{subequations}
\begin{align}
snr_o = \frac{|y|^2}{|\epsilon^y|^2} \\
snr_i = \frac{|x|^2}{|\epsilon^x|^2}  
\end{align}
\end{subequations}

We are interested in characterizing the factor $\alpha$ of reduction of the SNR through the inverse reconstruction:

\begin{equation}
\alpha = \frac{snr_i}{snr_o}
\end{equation}

To illustrate the mechanism through which the batch normalization inverse operation reduces the SNR, 
let us consider a toy layer with only two channels and parameters $\beta=[0,0]$ and $\gamma = [1, \rho]$. 
For simplicity, let us consider an input signal $x$ independently and identically distributed across both channels 
with zero mean and standard deviation 1 so that, in the forward pass, we have:

\begin{subequations}
\begin{align}
 y =& [y_0, y_1] \\
   =& [x_0, x_1 \times \rho] \\
|y|^2 =& |x_0|^2 + |x_1|^2 \times \rho^2 \\
      =& \frac{1}{2} \times |x|^2 + \frac{1}{2} \times |x|^2 \times \rho^2 \\
      =& \frac{|x|^2}{2} \times (1+\rho^2)
\end{align}
\end{subequations}

In which we used the assumption that $x$ is independently and identically distributed across both channels
to factorize $|x_0|^2 = |x_1|^2 = \frac{1}{2} \times |x|^2$ in equation (17d).
 
During the backward pass, the noisy estimate $\tilde{y}=y+\epsilon^y$ is fed back as input to the inverse operation. 
Similarly, let us suppose a noise $\epsilon^y$ identically distributed across both channels so that we have:

\begin{subequations}
\begin{align}
\tilde{y}       =& [ \tilde{y}_0, \tilde{y}_1 ] \\
                =& [ x_0 + \epsilon_0^y, x_1 \times \rho + \epsilon_1^y ] \\
\tilde{x}       =& [ \tilde{y}_0, \frac{\tilde{y}_1}{\rho}] \\
                =& [ x_0 + \epsilon_{0}^y, x_1 + \frac{\epsilon_{1}^y}{\rho} ]\\
\epsilon^x      =& \tilde{x} - x\\
                =& [ \epsilon_0^y, \frac{\epsilon_{1}^y}{\rho} ]\\
|\epsilon^x|^2  =& |\epsilon_0^y|^2 + \frac{|\epsilon_1^y|^2}{\rho^2} \\
                =& \frac{1}{2} \times |\epsilon^y|^2 + \frac{1}{2} \times \frac{|\epsilon^y|^2}{\rho^2} \\
                =& \frac{|\epsilon^y|^2}{2} \times (1 + \frac{1}{\rho^2})
\end{align}
\end{subequations}

Using the above formulation, the SNR reduction factor $\alpha$ can be expressed as:

\begin{subequations}
\begin{align}
\alpha =& \frac{snr_i}{snr_o} \\
 =& \frac{|x|^2}{|\epsilon^x|^2} \times  \frac{|\epsilon^y|^2}{|y|^2} \\
 =& \frac{4}{(1+\frac{1}{\rho^2}) \times (1 + \rho^2)}
\end{align}
\end{subequations}

Figure 5 shows the expected evolution of $\alpha$ through our toy layer for different values of the factor $\rho$.
To validate our formula, we empirically evaluate $\alpha$ for normal Gaussian inputs $x$ and output noise $\epsilon^y$ and find it to closely match the theoretical results given by equation 19. 

In essence, numerical instabilities in the inverse computation of the batch normalization layer arise
from the fact that the signal across different channels $i$ and $j$ are amplified by different factors $\gamma_i$ and $\gamma_j$.
While the signal amplification in the forward and inverse path cancel out each other ($x=f^{-1}(f(x))$),
the noise only gets amplified in the backward pass.

In the above demonstration, we have used a toy parameterization of the invertible batch normalization layer to illustrate the mechanism behind the SNR degradation. 
For arbitrarily parameterized batch normalization layers, the SNR degradation factor becomes:

\begin{subequations}
\begin{align}
\alpha =& \frac{snr_i}{snr_o} \\
       =& \frac{|x|^2}{|\epsilon^x|^2} \times  \frac{|\epsilon^y|^2}{|y|^2} \\
       =& \frac{|x|^2}{|y|^2} \times  \frac{|\epsilon^y|^2}{|\epsilon^x|^2} 
\end{align}
\end{subequations}

Assuming a noise $\epsilon^y$, equally distributed across all channels, the noise ratio can be computed as follows:

\begin{subequations}
\begin{align}
\tilde{y}_i  =& \gamma_i \times \frac{x_i - \hat{x_i}}{\sqrt{\dot{x_i}} + \epsilon} + \beta_i + \epsilon^y_i\\
\tilde{x}_i  =& (\sqrt{\dot{x_i}} + \epsilon) \times \frac{\tilde{y}_i -  \beta_i}{\gamma_i}  + \hat{x_i} \\
             =& x_i + \frac{\sqrt{\dot{x_i}}+\epsilon}{\gamma_i} \times \epsilon^y_i \\
\epsilon^x_i =& \tilde{x}_i - x_i\\
             =& \frac{\sqrt{\dot{x_i}}+\epsilon}{\gamma_i} \times \epsilon^y_i\\
\frac{|\epsilon^y|^2}{|\epsilon^x|^2} =& \frac{|\epsilon^y|^2}{\frac{|\epsilon^y|^2}{c} \times \sum_i \frac{\dot{x_i}^2}{\gamma_i^2}} \\
                                      =& \frac{c}{\sum_i \frac{\sqrt{\dot{x_i}}+\epsilon}{\gamma_i}}
\end{align}
\end{subequations}

Assuming input $x$ following a Gaussian distribution with channel-wise mean $\hat{x_i}$ and variance $\dot{x_i}$, 
the SNR reduction factor $\alpha$ becomes: 

\begin{subequations}
\begin{align}
\frac{|x|^2}{|y|^2} =& \frac{\sum_i |x_i|^2}{\sum_i|y_i|^2} \\
                    =& \frac{\sum_i (\hat{x}_i^2 + \dot{x_i})}{\sum_i (\gamma_i^2 + \beta_i^2)} \\
\alpha =& \frac{|x|^2}{|y|^2} \times  \frac{|\epsilon^y|^2}{|\epsilon^x|^2}  \\
       =& \frac{\sum_i (\hat{x}_i^2 + \dot{x_i})}{\sum_i (\gamma_i^2 + \beta_i^2)} \times \frac{c}{\sum_i \frac{\sqrt{\dot{x_i}}+\epsilon}{\gamma_i}}
\end{align}
\end{subequations}


Finally, we propose the following modification, introducing the hyperparameter $\epsilon_i$, to the invertible batch normalization layer:

\begin{subequations}
\begin{align}
y = f(x) &= |\gamma + \epsilon_i| \times \frac{x - \hat{x}}{\sqrt{\dot{x}} + \epsilon} + \beta \\
x = f^{-1}(y) &= (\sqrt{\dot{x}} + \epsilon) \times \frac{y -  \beta}{|\gamma + \epsilon_i|}  + \hat{x}
\end{align}
\end{subequations}

The introduction of the $\epsilon_i$ hyper parameter serves two purposes: 
First, it stabilizes the numerical errors described above by lower bounding the smallest $\gamma$ parameters. 
Second, it prevents numerical instabilities that would otherwise arise from the inverse computation as $\gamma$ parameters tend towards zero.

\subsubsection{Invertible activation function}

A good invertible activation function must be bijective (to guarantee the existence of an inverse function) and non-saturating (for numerical stability).
For these properties, we focus our attention on Leaky ReLUs whose forward $f$ and inverse $f^{-1}$ computations are defined, for a negative slope parameter $n$, as follow:

\begin{subequations}
\begin{align}
y = f(x) &=      \begin{cases}
x, & \text{if}\ x>0 \\
x / n, & \text{otherwise}
\end{cases} \\
x = f^{-1}(y) &= \begin{cases}
y, & \text{if}\ y>0 \\
y \times n, & \text{otherwise}
\end{cases} 
\end{align}
\end{subequations}

The analysis of the numerical errors yielded by the invertible Leaky ReLU follows a similar reasoning as the toy batch normalization example with an additional subtlety:
Similar to the toy batch normalization example, we can think of the leaky ReLU as artificially splitting the input x across two different channels, 
one channel leaving the output unchanged and one channel that divides the input by a factor $n$ during the forward pass and multiplies its output by a factor $n$ during the backward pass.

However, these artificial channels are defined by the sign of the input and output during the forward and backward pass respectively.
Hence, we need to consider the cases in which the noise flips the sign of the output activations, 
which leads to different behaviors of the invertible Leaky ReLU across four cases: 

\begin{subequations}
\begin{align}
y = \begin{cases}
y_{nn} \  \text{if}\  \hat{y}<0    &\text{and}\  y<0  \\
y_{np} \  \text{if}\  \hat{y}>=0   &\text{and}\  y<0  \\
y_{pp} \  \text{if}\  \hat{y}>=0   &\text{and}\  y>=0 \\
y_{pn} \  \text{if}\  \hat{y}<0    &\text{and}\  y>=1 
\end{cases} 
\end{align}
\end{subequations}

Where the index $np$, for instance, represents negative activations whose reconstructions have become positive due to the added noise. 
The signal to noise ratio of the input and outputs can be expressed respectively as:

In the case where $y >> \epsilon_y$, the probability of sign flips ($y_{np}$, $y_{pn}$) is negligible, 
so that the output signal $y$ is evenly split along $y_{pp}$ and $y_{nn}$.
In this regime, the degradation of the SNR obeys a formula similar to the toy batch normalization example:

\begin{subequations}
\begin{align}
 y =& [y_{pp}, y_{nn}] \\
   =& [x_{pp}, \frac{x_{nn}}{n}] \\
 |y|^2 =& \frac{1}{2} \times |x|^2 + \frac{1}{2} \times \frac{|x|^2}{n^2} \\
       =&\frac{|x|^2}{2} \times (1+\frac{1}{n^2})
\end{align}
\end{subequations}

\begin{subequations}
\begin{align}
\tilde{y}       =& [ \tilde{y}_{pp}, \tilde{y}_{nn}] \\
                =& [ x_{pp} + \epsilon_{pp}^y, \frac{x_{nn}}{n} + \epsilon_{nn}^y ] \\
\tilde{x}       =& [ \tilde{y}_{pp}, \tilde{y}_{nn} \times n] \\
                =& [ x_{pp} + \epsilon_{pp}^y, x_{nn} + \epsilon_{nn}^y \times n  ]\\
\epsilon^x      =& \tilde{x} - x\\
                =& [ \epsilon_{pp}^y, \epsilon_{nn}^y \times n ]\\
|\epsilon^x|^2  =& \frac{1}{2} \times |\epsilon^y|^2 + \frac{1}{2} \times |\epsilon^y|^2 \times n^2 \\
                =& \frac{|\epsilon^y|^2}{2} \times (1 + n^2)
\end{align}
\end{subequations}

Using the above formulation, the signal to noise ration reduction factor $\alpha$ can be expressed as:

\begin{subequations}
\begin{align}
\alpha =& \frac{snr_i}{snr_o} \\
       =& \frac{|x|^2}{|\epsilon^x|^2} \times  \frac{|\epsilon^y|^2}{|y|^2} \\
       =& \frac{4}{(1+\frac{1}{n^2}) \times (1 + n^2)}
\end{align}
\end{subequations}

Hence numerical errors can be controlled by setting the value of the negative slope $n$.
As $n$ tends towards $1$, $\alpha$ converges to $1$, yielding minimum signal degradation.
However, as $n$ tends towards $1$, the network tends toward a linear behavior, which hurts the model expressivity. 
Figure 6 shows the evolution of the SNR degradation $\alpha$ for different negative slopes $n$; 
and, in Section 5.1, we investigate the impact of the negative slope parameter on the model accuracy.

When the noise reaches an amplitude similar to or greater than the activation signal,
the effects of sign flips complicate the equation. 
However, in this regime, the signal to noise ratio becomes too low for training to converge,
as numerical errors prevent any useful weight update, so we leave the problem of characterizing this regime open.

\subsubsection{Invertible convolutions}

Invertible convolution layers can be defined in several ways.
The inverse operation of a convolution is often referred to as deconvolution,
and is defined for a subspace of the kernel weight space.

However, deconvolutions are computationally expensive and subject to numerical errors.
Instead, we choose to implement invertible convolutions using the channel partitioning scheme as the reversible block design for its simplicity, 
numerical stability and computational efficiency.
Hence, invertible convolutions, in our architecture, can be seen as minimal reversible blocks
 in which both modules consist of a single convolution.
Gomez \textit{et al.} \cite{gomez2017reversible} found the numerical errors introduced by reversible blocks to have no impact on the model accuracy. 
Similarly, we found reversible blocks extremely stable yielding negligible numerical errors
compared to the invertible batch normalization and Leaky ReLU layers.

\subsubsection{Pooling}

In \cite{jacobsen2018revnet}, the authors propose an invertible pooling operation that operates
by stacking the neighboring elements of the pooling regions along the channel dimension.
As noted in Section 3.5, the increase in channel size at each pooling level 
induces a quadratic increase in the number of parameters of upstream convolution, which creates a new memory bottleneck.

To circumvent this quadratic increase in the memory cost of the weight, 
we propose a new pooling layer that stacks the elements of neighboring pooling regions along the batch size instead of the channel size. 
We refer to both kind of pooling as channel pooling $\mathcal{P}_c$ and batch pooling $\mathcal{P}_b$ respectively, 
depending on the dimension along which activation features are stacked.
Given a $2 \times 2$ pooling region and an input activation tensor $x$ of dimensions $bs \times c \times h \times w$, 
where $bs$ refers to the batch size, $c$ to the number of channels and $h \times w$ to the spatial resolution, 
the reshaping operation performed by both pooling layers can be formalized as follows:

\begin{subequations}
\begin{align}
	\mathcal{P}_c :& x \rightarrow y \\
	              :&  \mathbb{R}^{bs \times c \times h \times w}  \rightarrow \mathbb{R}^{bs \times 4c \times \frac{h}{2} \times \frac{w}{2}}\\
	\mathcal{P}_b :& x \rightarrow y \\
                      :&  \mathbb{R}^{bs \times c \times h \times w}  \rightarrow \mathbb{R}^{4bs \times c \times \frac{h}{2} \times \frac{w}{2}}
\end{align}
\end{subequations}

Channel pooling gives us a way to perform volume-preserving pooling operations while increasing the number of channels at a given layer of the architecture,
while batch pooling gives us a way to perform volume-preserving pooling operations while keeping the number of channel constant, 
By alternating between channel and batch pooling, we can control the number of channels at each pooling level of the model's architecture.

\subsubsection{Layer-wise invertible architecture}

Putting together the above building blocks, Figure 7 illustrates a layer-wise invertible architecture.
The peak memory usage for a training iteration of this architecture, as parameterized in Table 1, can be computed as follows:

\begin{subequations}
\begin{align}
\mathcal{M} &= M_{\theta} + M_{z} + M_{g} \\
            &= M_{\theta} + (M_z' + M_{g}') \times (h \times w \times bs) \\
            &= 29.6 \times 10^6 + 320 \times (h \times w \times bs)
\end{align}
\end{subequations}

Training an iteration over a typical batch of 32 images with resolution $240 \times 240$ would require $\mathcal{M}=590$MB of VRAM. 
Similar to the RevNet architecture, the reconstruction of the hidden activations by inverse transformations during the backward pass comes with an additional computational cost similar to a forward pass.

\subsection{Hybrid architecture}

In section 3, we saw that layer-wise activation and normalization layers degrade the signal to noise ratio of the reconstructed activations.
In section 5.1, we will quantify the accumulation of numerical errors through long chains of layer-wise invertible operations and show that numerical errors negatively impact model accuracy.

To prevent these numerical instabilities, we introduce a hybrid architecture, illustrated in Figure 8, combining reversible residual blocks with layer-wise invertible functions.
Conceptually, the role of the residual level reversible block is to reconstruct the input activation of residual blocks with minimal errors, 
while the role of the layer-wise invertible layers is to efficiently recompute the hidden activations within the reversible residual 
blocks at the same time as the gradient propagates to circumvent the local memory bottleneck of the reversible module.

The backward pass through these hybrid reversible blocks is illustrated in Figure 9 and proceeds as follows: 
First, the input $x$ is computed from the output $y$ through the analytical inverse of the reversible block.
These computations are made without storing the hidden activation values of the sub-modules.
Second, the gradient of the activations are propagated backward through the reversible of the block modules.
As each layer within these modules is invertible, the hidden activation values
are computed using the layer-wise inverse along the gradient.

The analytical inverse of the residual level reversible blocks is used to propagate hidden activations with minimal reconstruction error to the lower modules,
while layer-wise inversion allows us to alleviate the local bottleneck of the reversible block by computing the hidden activation values together with the backward flow of the gradients. 
As layer-wise inverses are only used for hidden feature computations within the scope of the reversible block,
and reversible blocks are made of relatively short chains of operations,
numerical errors do not accumulate up to a damaging degree.

The peak memory consumption of our proposed architecture, as illustrated in Figure 8 and parameterized in Table 1, can be computed as  
\begin{subequations}
\begin{align}
\mathcal{M} &= M_{\theta} + M_{z} + M_{g} \\
            &= M_{\theta} + (M_z' + M_{g}') \times (h \times w \times bs) \\
            &= 14.8 \times 10^6 + 352 \times (h \times w \times bs) 
\end{align}
\end{subequations}

Training an iteration over batch of 32 images of resolution $240 \times 240$ would require $\mathcal{M}=648$MB of VRAM.

It should be noted, however, that this architecture adds an extra computational cost as both the reversible block inverse and layer-wise inverse need to be computed.
Hence, instead of one additional forward pass, as in the RevNet and layer-wise architectures,
our hybrid architecture comes with a computational cost equivalent to performing two additional forward passes
during the backward pass. 

\section{Results and Discussion}

We use the CIFAR10 dataset as a benchmark for our experiments.
The CIFAR10 dataset is complex enough to require efficient architectures to reach high accuracy, 
yet small enough to enable us to rapidly iterate over different architectural designs.
We start by analyzing numerical errors arising in layer-wise invertible and hybrid architectures, 
and outline their impact on accuracy.
This analysis motivates our choice of architecture and hyperparameter.
We then summarize the benefits and drawbacks of our proposed architecture in comparison to different baseline architectures.

\subsection{Impact of Numerical stability}

\subsubsection{Layer-wise Invertible Architecture}

In this section, we quantify the accumulation of numerical errors in layer-wise invertible architectures and analyze their impact on the accuracy.
The architecture of these models is illustrated in Figure 7.
We investigate the evolution of numerical errors, and their impact on accuracy, for networks of different depth and different hyper-parameter values.
Figure 10 illustrates the degradation of the signal-to-noise ration along the layers of one such model.

We found the two most impacting parameters to be the depth $N$ of the network and the negative slope $n$ of the activation function.
Figure 11 shows the evolution of the numerical errors with both of these parameters.

Next, we investigate the impact of numerical errors on the accuracy.
In order to isolate the impact of the numerical errors, 
we compare the accuracy reached by the same architecture with and without inverse reconstruction of the hidden layers activations.
Without reconstruction, the hidden activation values are stored along the forward pass and the gradient updates are computed from the true, 
noiseless activation values, so that the only difference between both settings is the noise introduced by the inverse reconstructions.

In Figure 12, we compare the evolution of the accuracy in both settings for different depth and negative slopes.
For small depths (or high negative slopes), in which the numerical errors are minimum, both models yield similar accuracy.
However, as the numerical errors grow, the accuracy of the model goes down, 
while the accuracy of the ideal baseline keeps increasing,
which can be seen with both depth and negative slopes.
This loss in accuracy is the direct result of numerical errors, 
which prevent the model from converging to higher accuracies.

\subsubsection{Hybrid Invertible Architecture}

In section 4.2, we introduced a hybrid architecture,
illustrated in Figure 8, to prevent the impact of numerical errors on accuracy.
Figure 13 shows the propagation of the signal to noise ratio through the layers of such hybrid architecture.
As can be seen in this figure, the hybrid architecture is much more robust to numerical errors as activations 
are propagated from one reversible block to the other using the reversible block inverse computations instead of layer-wise inversions.

Figure 14 shows the evolution of the SNR with increasing depth $N$ and for different values of negative slope $n$.
This figure shows a much more stable evolution of the signal to noise ratio than the layer-wise architecture. 

Figure 15 compares the evolution of the accuracy reached by this hybrid architecture with noisy activations and noiseless ideal activations as depth and negative slope increase.
The negative impacts of numerical errors observed in the layer-wise architecture are gone,
confirming that the numerical stability brought by the hybrid architecture effectively stabilizes training.

\subsection{Model comparison}

Table 1 summarizes our main results.
In this table, we compare architectures with different patterns of reversibility.
To allow for a fair comparison, we have tweaked each architecture to keep the number of parameters as close as possible,
with the notable exception of the i-RevNet architecture. 
The i-Revnet pooling scheme enforces a quadratic growth of its parameters with each level of pooling.
In order to keep the number of parameters of the i-RevNet close to the other baselines, we would have to drastically 
reduce the number of channels of lower layers, which we found yield poor performance. 
Furthermore, it should be noted that the i-RevNet architecture we present slightly differs from the original i-Revnet model
as our implementation uses RevNet-like reversible modules with one module per channel split for similarity with
the other architecture we evaluate instead of the single module used in the original architecture.

All models were trained for 50 epochs of stochastic gradient descent with cyclical learning rate and momentum \cite{smith2017super} with minimal image augmentation.

\begin{table*}[t]
\begin{tabular}{ c c c c c c c c}	
Model     & Accuracy & \#Params & Channels & Pooling  & $M_{\theta}$ & $M_{z}'+M_{g}'$ & $\mathcal{M} $ \\
\hline
Resnet    & $94.7\%$   & $3.1M$   &  $32 - 64 - 128  - 256$       & Max Pooling      				           &  $12.5M$   &  $1928$  & $1.01G$  \\			
RevNet    & $94.5\%$   & $3.1M$   &  $40 - 80 - 256  - 320$       & Max Pooling      				           &  $12.7M$   &  $640$   & $348M$   \\
i-RevNet  & $93.8\%$   & $42.8M$  &  $32 - 128 - 512 - 2048$      & $\mathcal{P}_c - \mathcal{P}_c - \mathcal{P}_c$          &  $171M$    &  $640$   & $500M$   \\
Ours      & $93.3\%$   & $3.7M$   &  $32 - 128 - 512 - 512$       & $[\mathcal{P}_c, \mathcal{P}_c, \mathcal{P}_b]$          &  $14.8M$   &  $352$   & $200M$   \\
\hline
\end{tabular}
\begin{center}
\caption{Summary of architectures with different levels of reversibility}
\end{center}
\end{table*}

The parameters of our proposed architecture are given in Table 1.
This architecture was selected as the best performing architecture 
from an extensive architecture search on a constrained weight budget. 
Compared to the original ResNet architecture, our model drastically cuts the memory cost of training.
These drastic memory cuts come at the cost of a small degradation in accuracy.

\begin{table}[h]
\centering
\begin{tabular}{ c c c c}	
 GPU & Accuracy  & Time \\
\hline			
GTX750     & $93.3\%$  & $35 min.$    \\
GTX 1080Ti & $93.3\%$  & $67 min.$  \\
\hline
\end{tabular}
\caption{Training statistics on different hardware}
\end{table}

Furthermore, our hybrid architecture requires the computational equivalent of two additional forward passes within each backward pass.
The computational complexity, however, remains reasonable:
In Table 2, we compare the time of training our proposed architecture to 93.3\% on a high-end Nvidia GTX 1080Ti and a low-end
Nvidia GTX750.
The GTX750 only has 1GB of VRAM, which results in roughly 400MB of available memory after the initialization of various frameworks.
Training a vanilla ResNet with large batch sizes on such limited memory resources is impractical, while our architecture allows for efficient training. 

\section{Conclusion}

Convolutional Neural Networks form the backbone of modern computer vision systems.
However, the accuracy of these models come at the cost of resource intensive training and inference procedures.
While tremendous efforts have been put into the optimization of the inference step on resource-limited device,
relatively little work have focused on algorithmic solutions for limited resource training. 
In this paper, we have presented an architecture able to yield high accuracy classifications within very tight memory constraints.
We highlighted several potential applications of memory-efficient training procedures, such as on-device training,
and illustrated the efficiency of our approach by training a CNN to 93.3\% accuracy on a low-end GPU with only 1GB of memory.

\section{FIGURE LEGEND}

\begin{figure}[t]
\includegraphics[width=0.9\linewidth]{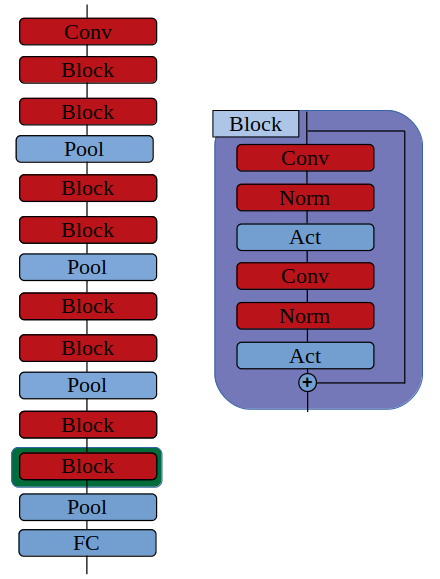}
\caption{
Illustration of the ResNet-18 architecture and its memory requirements.
Modules contributing to the peak memory consumption are shown in red.
These modules contribute to the memory cost by storing their input in memory.
The green annotation represents the extra memory cost of storing the gradient in memory.
The peak memory consumption happens in the backward pass through the last convolution so that this layer is annotated with an additional gradient memory cost.
At this step of the computation, all lower parameterized layers have stored their input in memory, which constitutes the memory bottleneck.  
}
\end{figure}

\begin{figure}[t]
\includegraphics[width=0.5 \textwidth]{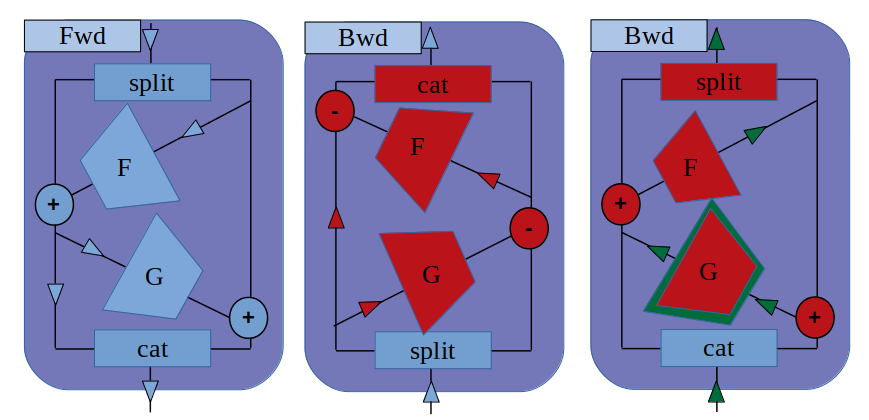}
\caption{
Illustration of the backpropagation process through a reversible block.
In the forward pass (left), activations are propagated forward from top to bottom.
The activations are not kept in live memory as they are to be recomputed in the backward pass so no memory bottleneck occurs.
The backward pass is made of two phases: 
First the hidden and input activations are recomputed from the output through an additional forward pass through both modules (middle).
Once the activations recomputed, the activations gradient are propagated backward through both modules of the reversible blocks (right).
Because the activation and gradient computations flow in opposite directions through both modules,
both computations cannot be efficiently overlapped, which results in the local memory bottleneck of 
storing all hidden activations within the reversible block before the gradient backpropagation step.
}
\end{figure}

\begin{figure}[t]
\includegraphics[width=0.5 \textwidth]{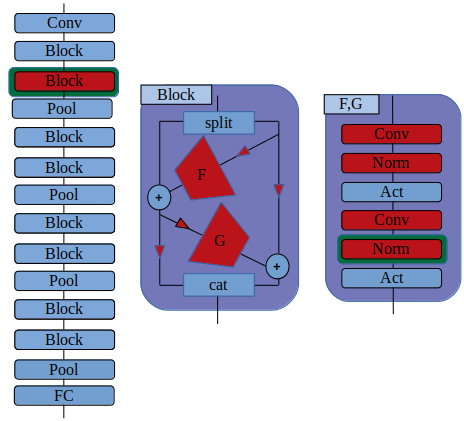}
\caption{
Illustration of the Revnet architecture and its memory consumption.
Modules contributing to the peak memory consumption are shown in red.
The peak memory consumption happens during the backward pass through the first reversible block.
At this step of the computations, all hidden activations within the reversible block are stored in memory simultaneously.
}
\end{figure}

\begin{figure}[t]
\includegraphics[width=0.5\textwidth]{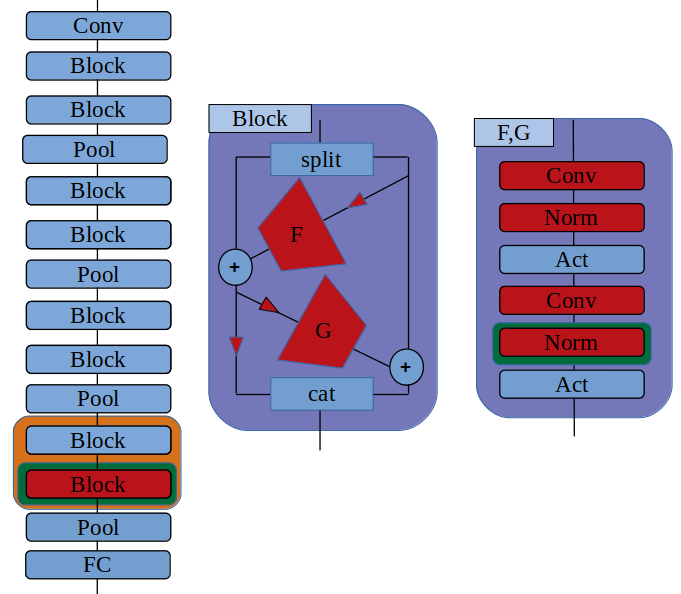}
\caption{
Illustration of the i-Revnet architecture and its memory consumption.
The peak memory consumption happens during the backward pass through the top reversible block.
In addition to this local memory bottleneck, the cost of storing the top layers weights 
(in orange) becomes a new memory bottleneck as the 
weight kernel size grows quadratically in the number of channels.
}
\end{figure}

\begin{figure}[t]
\includegraphics[width=0.9\linewidth]{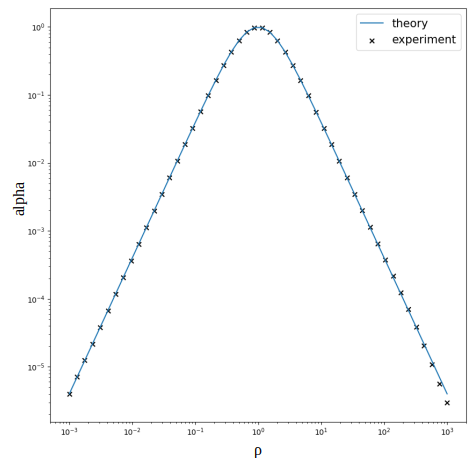}
\caption{
Illustration of the numerical errors arising from batch normalization layers.
Comparison of the theoretical and empirical evolution of the $\alpha$ ratio for different $\rho$ values in our toy example.
Empirical values were computed for a Gaussian input signal with zero mean and standard deviation 1 and a white Gaussian noise of standard deviation $10^{-5}$.
}
\end{figure}

\begin{figure}[t]
\includegraphics[width=0.9\linewidth]{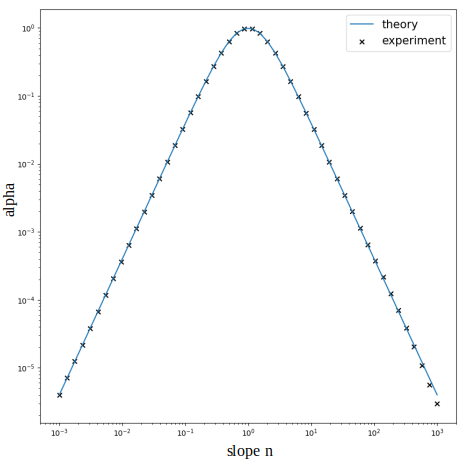}
\caption{
Illustration of the numerical errors arising from invertible activation layers.
Comparison of the theoretical and empirical evolution of the $\alpha$ ratio for different negative slopes $n$.
Empirical values were computed for a Gaussian input signal with zero mean and standard deviation 1 and a white Gaussian noise of standard deviation $10^{-5}$.
}
\end{figure}
\begin{figure}[t]
\includegraphics[width=0.9\linewidth]{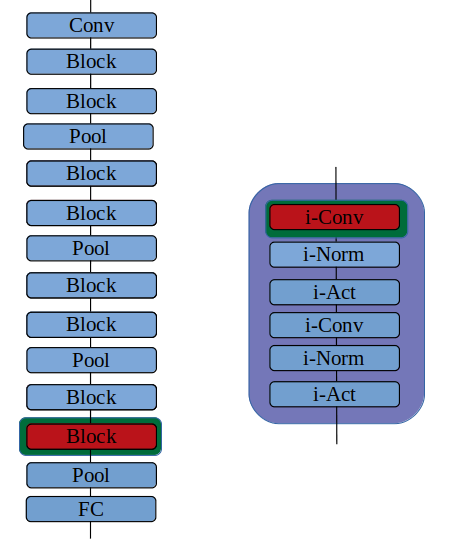}
\caption{
Illustration of a layer-wise invertible architecture and its memory consumption. 
}
\end{figure}

\begin{figure}[t]
\includegraphics[width=0.9\linewidth]{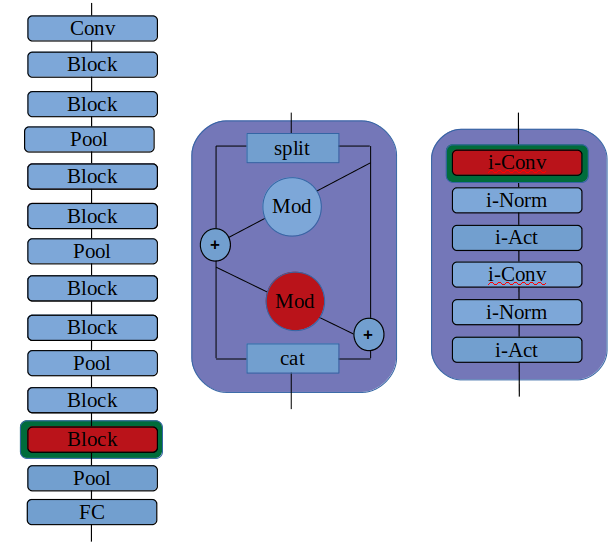}
\caption{
Illustration of a hybrid architecture and its peak memory consumption. 
}
\end{figure}

\begin{figure}[t]
\includegraphics[width=0.9\linewidth]{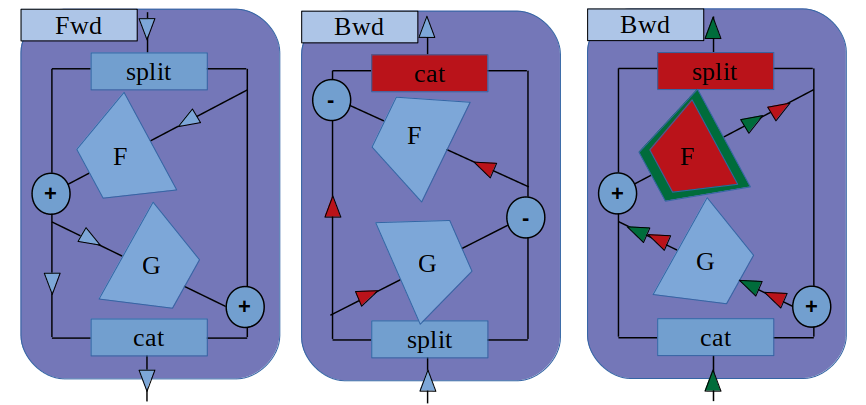}
\caption{
Illustration of the backpropagation process through a reversible block of our proposed hybrid architecture.
In the forward pass (left), activations are propagated forward from top to bottom.
The activations are not kept in live memory as they are to be recomputed in the backward pass so that no memory bottleneck occurs.
The backward pass is made of two phases: 
First the input activations are recomputed from the output using the Reversible block analytical inverse (middle).
This step allows to reconstruct the input activations with minimal reconstruction error.
During this step, hidden activations are not kept in live memory so as to avoid the local memory bottleneck of the reversible block.
Once the input activation recomputed, the gradients are propagated backward through both modules of the reversible blocks (right).
During this second phase, hidden activations are recomputed backward through each module using the layer-wise inverse operations, yielding minimal memory footprint
}
\end{figure}

\begin{figure}[t]
\includegraphics[width=0.9\linewidth]{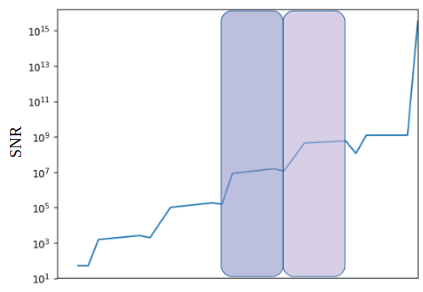}
\caption{
Evolution of the SNR through the layers of a layer-wise invertible model.
Color boxes illustrate the span of two consecutive convolutional blocks (Convolution-normalization-activation layers).
The SNR gets continuously degraded throughout each block of the network, resulting in numerical instabilities.
}
\end{figure}

\begin{figure}[t]
\includegraphics[width=0.9\linewidth]{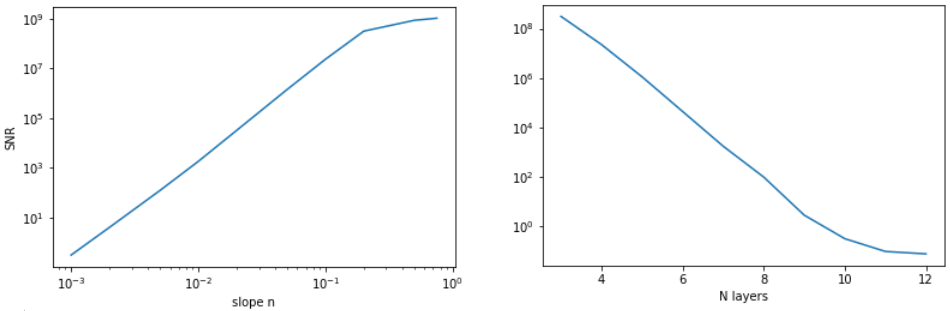}
\caption{
Illustration of the impact of depth (in number of layers $N$) and negative slope $n$ on the numerical errors.
Both figure shows the evolution of the SNR at the lowest layer of a layer-wise invertible network with increasing depth and negative slopes.
The lower the SNR is, the more  important numerical errors of the inverse reconstructions are. 
(Left): The SNR decreases exponentially with depth until it reaches an SNR value of 1. 
At this point, the noise is of the same scale as the signal, and no learning can happen.
These results were computed with a negative slope of $n=2$
(Right) This figure shows the evolution of the SNR with different negative slopes $n$ for a layer-wise reversible model of depth 3.
On a log-log scale, this figure shows an almost linear relationship between negative slope and SNR.
It is impressive that with only three layer depth, a negative slope of $n=10^{-3}$ reaches a SNR superior to 1.
With such parameterization, even the most shallow models are not capable of learning.
}
\end{figure}

\begin{figure}[t]
\includegraphics[width=0.9\linewidth]{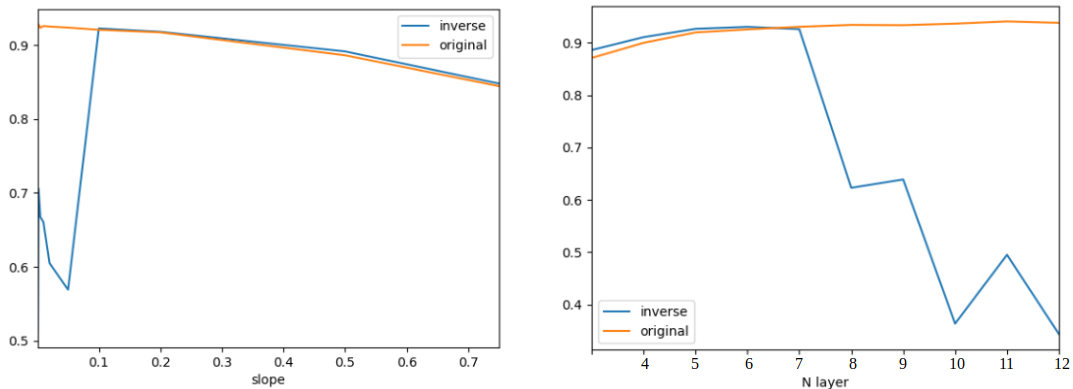}
\caption{
Impact of the numerical errors on the accuracy of layer-wise invertible models.
(Left): Evolution of a 6-layer model accuracy with and without inverse reconstructions with the negative slope.
Without reconstruction, the model accuracy benefits from smaller negative slopes.
With inverse reconstructions, the model similarly benefits from smaller negative slopes as $n$ decreases from $1$ to $0.1$.
For smaller negative slopes, however, the accuracy sharply decreases toward lower values due to numerical errors.
(Right) Evolution of the accuracy with depth for a negative slope $n=0.2$ with and without inverse reconstructions.
Without reconstruction, the model accuracy benefits from depth.
With inverse reconstructions, the model similarly benefits from depth as the number of layers grow from $3$ to $7$.
For $N>7$, however, the accuracy sharply decreases toward lower values due to numerical errors.
}
\end{figure}

\begin{figure}[t]
\includegraphics[width=0.9\linewidth]{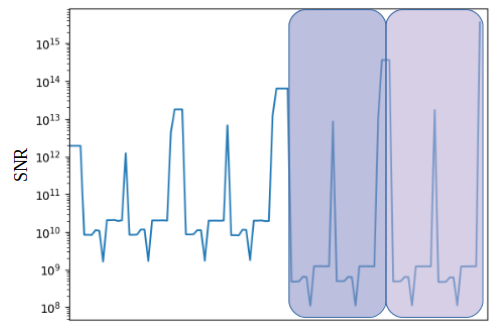}
\caption{
Evolution of the SNR through the layers of a hybrid architecture model.
The span of two consecutive reversible blocks are shown with color boxes.
Within reversible blocks, the SNR quickly degrades due to the numerical errors introduced by invertible layers.
However, the signal propagated to the input of each reversible block is recomputed using the reversible block inverse, which is much more stable.
Hence, we can see a sharp decline of the SNR within the reversible blocks, but the SNR almost raises back to its original level at the input of each reversible block. 
}
\end{figure}

\begin{figure}[t]
\includegraphics[width=0.9\linewidth]{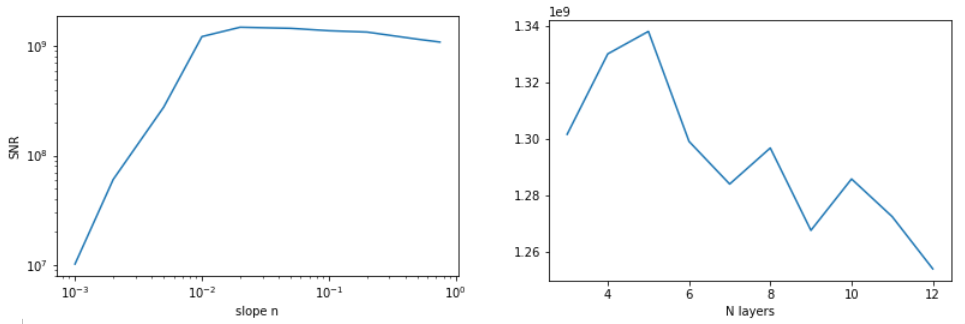}
\caption{
Illustration of the impact of depth (in number of layers $N$) and negative slope $n$ on the numerical errors.
Both figure shows the evolution of the SNR at the lowest layer of our hybrid architecture with increasing depth and negative slopes.
Our hybrid architecture greatly reduce the impact of both depth and negative slopes on the numerical errors
}
\end{figure}

\begin{figure}[t]
\includegraphics[width=0.9\linewidth]{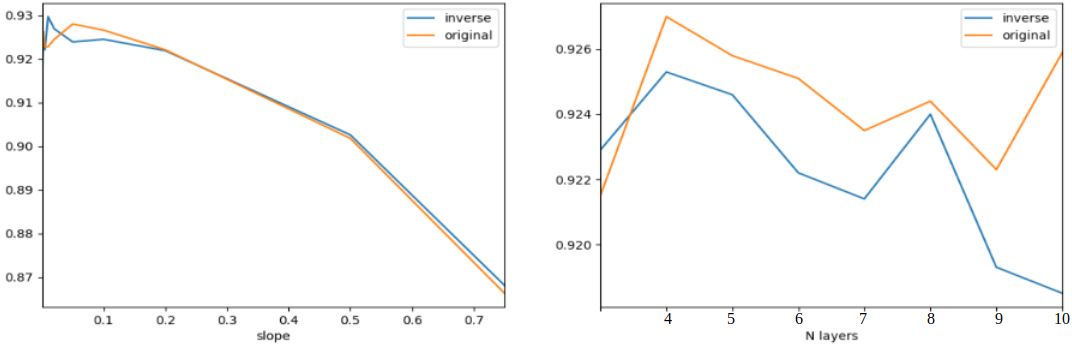}
\caption{
Impact of the numerical errors on the accuracy of layer-wise invertible models.
Our proposed hybrid architecture greatly stabilizes the numerical errors,
which results in smaller effects of the depth and negative slope on accuracy. 
}
\end{figure}

\bibliographystyle{bmc-mathphys}
\bibliography{bmc_article}

\end{document}